\documentclass[letterpaper,journal,10pt]{IEEEtran}
\usepackage[utf8]{inputenc}
\usepackage{amsmath,amsfonts,amssymb}
\usepackage{graphicx}
\usepackage{cite}
\usepackage{booktabs} 
\usepackage{xcolor}   
\usepackage{siunitx}
\usepackage{tabularx}
\usepackage{algorithm}
\usepackage[noend]{algpseudocode} 
\usepackage[hidelinks]{hyperref} 
\usepackage{multirow}

\usepackage{float}
\begin{document}

\title{Scene-Adaptive Continual Learning for CSI-based Human Activity Recognition with Mixture of Experts}

\author{%
  Wenhan Zheng, 
  Yuyi Mao,~\IEEEmembership{Senior Member,~IEEE}, 
  Ivan Wang‐Hei Ho,~\IEEEmembership{Senior Member,~IEEE}
  
  \thanks{%
    W.~Zheng and I.~W.~Ho are with the Department of Electrical and Electronic Engineering,
    The Hong Kong Polytechnic University, Hong Kong (e-mail:
    kevin-wenhan.zheng@connect.polyu.hk; ivanwh.ho@polyu.edu.hk).Y.~Mao is with the School of Computer Science and Engineering,
    The Macau University of Science and Technology, Macau (e-mail:
    yymao@must.edu.mo)%
  }
  \thanks{
        \textit{(Corresponding authors: Ivan Wang-Hei Ho and Yuyi Mao).}
    }
}

\maketitle

\begin{abstract}
Channel state information (CSI)-based human activity recognition (HAR) is vulnerable to performance degradation under domain shifts across varying physical environments. Continual learning (CL) offers a principled way to learn new domains sequentially while preserving past knowledge, but existing CL solutions for CSI-based HAR scale poorly with accumulating domains, rely on a large replay buffer, or incur linearly growing inference cost. In this letter, we propose Scene-Adaptive Mixture of Experts with Clustered Specialists (SAMoE-C), which formulates cross-domain CSI-based HAR as a mixture-of-experts system that enables scene-specific adaptation, via an attention-based semantic router that activates only selected experts for each input. Moreover, we develop a novel training protocol, which requires only a tiny replay buffer for stabilizing domain discrimination of the router. Experimental results on a four-scene CSI dataset demonstrate that SAMoE-C approaches the state-of-the-art accuracy, while maintaining a significantly lower inference cost. By jointly combining modular experts, selective activation with router and a lightweight training protocol, SAMoE-C enables scalable cross-domain CSI-based HAR deployment with low training overhead and high computational efficiency in real-world settings.
\end{abstract}


\section{Introduction}
\label{sec:intro}
\IEEEPARstart{C}{hannel} state information (CSI) is a fine-grained physical layer measurement within wireless communication networks that grants a unique opportunity with privacy-preserving human activity recognition (HAR)~\cite{10.1145/3530682}. The high sensitivity of CSI to the propagation environment, characterized by multipath fading, enables the extraction of human motion-related features. However, unlike visual data such as images, CSI signals do not retain object structure across domains. As a result, the accuracy of a pre-trained model drops notably when deployed in a new physical scene due to domain shift, as the same activity can lead to substantially different signatures in CSI~\cite{10836707}.

The domain shift challenge of HAR was primarily addressed via transfer learning and domain adaptation techniques~\cite{dhekane2024transfer, cook2013transfer}. These methods align feature distributions between a labeled source domain and an unlabeled target domain, often by minimizing statistical distances or using adversarial training to learn domain-invariant representations~\cite{ganin2016domain}. While effective, these strategies assume concurrent access to data from both source and target domains. This assumption is often impractical because CSI data of historical domains may not be fully retrievable due to storage or privacy constraints.

A robust CSI-based HAR system must continuously learn from new environments without retraining the deep neural network (DNN) models from scratch. This brings a core challenge named catastrophic forgetting, i.e., knowledge of previous domains learned by DNNs may be overwritten when they are adapting to new ones. Continual learning (CL) techniques can address this challenge through three typical approaches: 1) To protect weights learned from data of past domains by regularizing the loss function~\cite{10.5555/3692070.3694593}; 2) To maintain a replay buffer of historical data for domain adaptation~\cite{rebuffi2017icarl}; and 3) To dynamically expand the network architecture to isolate parameters related to specific domains~\cite{zhou2023model}. 

However, these CL approaches face limitations in CSI-based HAR. Firstly, regularization-based methods rely on DNN models with a fixed set of predefined trainable parameters. As more domains appear progressively, performance of regularization-based models may eventually be degraded. Secondly, replay mechanisms that rehearse decision boundaries for many activities are sensitive to the replay buffer size, showing inferior performance on resource-constrained edge devices, which can only cache a limited amount of historical data. Although existing adaptive DNN architectures, such as MEMO~\cite{zhou2023model} expand by adding modules for new domains to avoid performance saturation, all module outputs are aggregated for inference, causing linearly growing computational complexity.

To learn CSI-based HAR from multiple domains emerging in sequence, we utilize the Mixture of Experts (MoE)~\cite{shazeer2017outrageously}, which selectively activates a subset of specialized expert networks according to input data, while maintaining low computation cost regardless of the number of domains. We propose the Scene-Adaptive Mixture of Experts with Clustered Specialists (SAMoE-C) framework and our main contributions are summarized as follows: We cast the cross-domain CSI-based HAR problem into a CL formulation. To enable efficient domain adaptation, we develop a DNN model based on a MoE architecture that activates domain-specific expert networks guided by a semantic router. Moreover, rather than extensively replaying CSI data for class-level domain adaptation at experts, we propose an efficient training protocol that requires only a tiny replay buffer set for training the router to stabilize domain discrimination. Empirical experiments demonstrate a good balance between accuracy and efficiency of SAMoE-C compared to the state-of-the-art methods.

\section{System Model and Problem Formulation}
\label{sec:model}

\subsection{CSI Data Model}
CSI can be obtained from commodity WiFi devices operating with Multiple-Input Multiple-Output (MIMO) Orthogonal Frequency Division Multiplexing (OFDM) technologies ~\cite{bolcskei2006mimo}. For a single packet transmission at time $t$, the CSI across $S$ OFDM subcarriers can be described by the channel frequency response (CFR) as follows:
\begin{equation}
\label{eq:cfr}
\mathbf{H}(t) = [H(f_1, t), H(f_2, t), \dots, H(f_S, t)],
\end{equation}
where $H(f_k, t) \in \mathbb{C}^{N_{rx} \times N_{tx}}$ is the complex-valued channel matrix for the $k$-th subcarrier, with $N_{tx}$ and $N_{rx}$ respectively being the numbers of transmit and receive antennas. Human motion within the environment induces fluctuations in the signal propagation paths, which are captured as variations in the CSI. We utilize the amplitude of the CSI for HAR. A single activity instance is captured over a window of $T$ time instances, forming a data sample $\mathbf{X} \in \mathbb{R}^{C \times T \times S}$, where $C \triangleq N_{tx} \times N_{rx}$. This tensor representation effectively captures the spatial-temporal variations of the CSI signals that are valuable for HAR~\cite{10476327}.

\subsection{Cross-Domain CSI-Based HAR Problem}
We consider a set of $K$ distinct human activities, with the corresponding label space denoted as $\mathcal{Y} = \{y_0, y_1, \dots, y_{K-1}\}$. The main difficulty arises from the change of physical environment, referred to as domain shift. Even though the label space $\mathcal{Y}$ is shared across environments, the CSI distribution is dependent on domain-specific multipath fading patterns.

Specifically, we consider a sequence of $N$ domains. For each domain $e \in \{1, \dots, N\}$, we collect a dataset $\mathcal{D}_e = \{(\mathbf{X}_i, y_i)\}_{i=1}^{|\mathcal{D}_e|}$, where $\mathbf{X}_i$ is the CSI tensor and $y_i \in \mathcal{Y}$ is the activity label. The conditional probability distributions of the CSI data $\mathbf{X}$ given label $y$ vary across domains, i.e., for any two distinct domains $e \neq e'$, there exists $\mathbf{X}$ such that:
\begin{equation}
\label{eq:domain_shift}
P_e(\mathbf{X}|y) \neq P_{e'}(\mathbf{X}|y).
\end{equation}
Consequently, a model trained solely on data from domain $\mathcal{D}_e$ may exhibit reduced generalization when deployed in a different domain $\mathcal{D}_{e'}$ due to domain shift. This necessitates a domain-adaptive approach that learns new domains while retaining previously acquired knowledge.

\begin{figure}[t]
    \centering
    \includegraphics[width=\columnwidth]{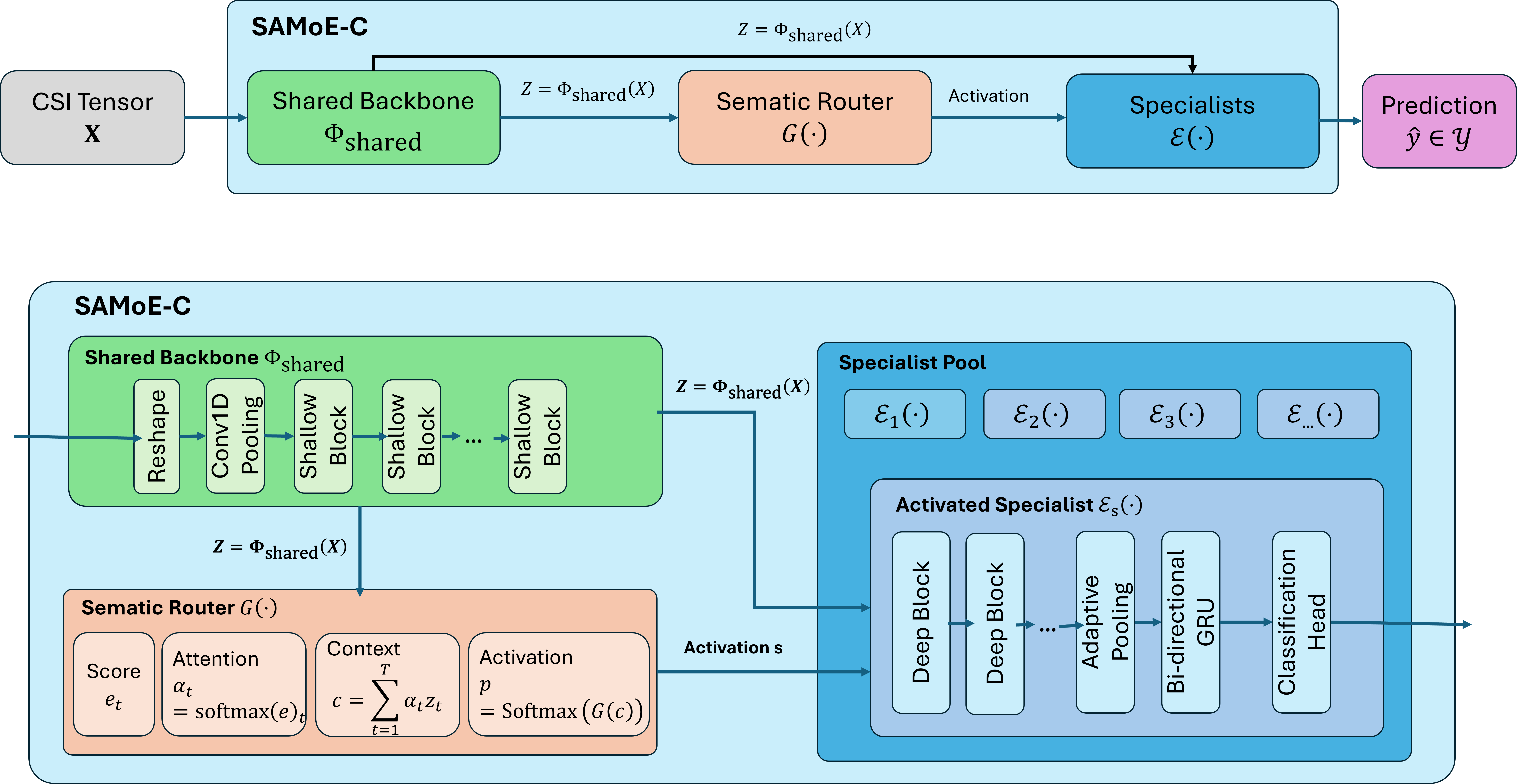}\caption{The overall architecture of our proposed SAMoE-C framework. The framework features a modular data processing pipeline with multiple specialists. }
    \label{fig:framework}
\end{figure}

\subsection{CL Formulation}
We formulate the cross-domain CSI-based HAR problem as a Continual Learning (CL) process. Let's denote $F_\theta$ as a deep neural network model parameterized by $\theta$ that maps CSI tensors to activity labels. The model is to be optimized over a sequence of $N$ CSI-based HAR tasks, denoted as $\mathcal{T} = \{\mathcal{T}_1, \mathcal{T}_2, \dots, \mathcal{T}_N\}$, where each task corresponds to a different physical environment. A \textit{learning step} $s$ is formally defined as the adaptation phase where the classification model updates its parameters using only the dataset $\mathcal{D}_s$ collected in the $s$-th environment. This constraint is driven by the limited storage quota on edge devices and security-aware data retention policy. However, a small fraction $\rho$ of the CSI data in each domain is typically retained in a cache buffer for control purposes and channel estimation. The learning objective is to minimize the expected classification loss via the following function:
\begin{equation}
\label{eq:learnobj}
    \min_{\theta} \sum_{s=1}^{N} \mathbb{E}_{(\mathbf{X}, y) \sim P_s} [\mathcal{L}(F_\theta(\mathbf{X}), y)],
\end{equation}
where $P_s$ denotes the CSI data distribution induced by the $s$-th propagation environment.

Optimizing (\ref{eq:learnobj}) presents significant challenges. On one hand, unlike vision-based sensing where domain-invariant properties can be easily transferred, the high sensitivity of CSI to environments could lead to rapidly changing multipath patterns, which can cause catastrophic forgetting across domains. On the other hand, $N$ can be large, which exceeds the capacity of monolithic models with a fixed number of trainable parameters. Although architecture-expansion strategies, e.g., the class incremental expansion introduced by MEMO model, propose a solution to the limited model capacity with fixed trainable parameters, it is not suitable for domain incremental problems with fixed classes, and aggregating all modules at inference time incurs linearly increasing overhead~\cite{zhou2023model}. Our goal is to propose a modular expansion-based model tailored to domain incremental problems and maintain accuracy in future domains without unbounded inference-cost growth.

\section{The Proposed SAMoE-C Framework}
\label{sec:framework} 

To address the challenges of cross-domain CSI-based HAR, we propose the SAMoE-C framework, as shown in Fig.~\ref{fig:framework}. The framework is designed to separate the network architecture into two modules, where a shallow \textbf{shared feature extraction backbone} collaborates with deeper \textbf{activity classification specialist networks}. As such, the activity classification networks can be swapped \textit{per domain} to balance the stability of feature extraction and scalability of classification. Specifically:

\begin{itemize}
    \item The shared feature extraction backbone, denoted as $\Phi_{\text{shared}}\left(\cdot\right)$, transforms the raw CSI tensor $\mathbf{X}$ into an intermediate representation through a multi-layer DNN model. The output is denoted as a sequence of latent feature vectors capturing temporal CSI signal variations. 
    \item The output from the shared feature extraction backbone is passed to the activity classification specialists, which are a set of independent DNNs denoted as ${E_{k}\left(\cdot\right)}$. Each specialist network is optimized for a specific domain and consists of \textit{Specialist Block} which has deeper DNN layers and a linear \textit{Softmax Head} that determines the class of human activity. 
\end{itemize}

\label{subsec:gating}
One novelty of SAMoE-C framework is sparse activation of specialist blocks according to a \textbf{semantic router}, which is guided by a \textbf{temporal attention mechanism}. The router distinguishes different domain patterns to select the most appropriate processing pathway. In the following, we detail the processing in the temporal attention mechanism and semantic router, while the training procedures of the SAMoE-C framework will be discussed in Section 
\ref{sec:incremental_learning}. 

\subsection{Temporal Attention}
To handle the duration variability of human activities, we employ a standard Additive Attention mechanism. Let $\mathbf{Z} \in \mathbb{R}^{T \times d}$ represent the sequence of spatially global-averaged feature vectors output by the shared backbone. We first compute a score for each time step $t$ through a 2-layer network as follows:
\begin{equation}
    e_t = \mathbf{v}^\top \tanh(\mathbf{W} \mathbf{z}_t + \mathbf{b}),
\end{equation}
where $\mathbf{W} \in \mathbb{R}^{d_h \times d}$, $\mathbf{b} \in \mathbb{R}^{d_h}$, and $\mathbf{v} \in \mathbb{R}^{d_h}$ are learnable parameters. We then calculate the attention weights $\alpha_t$ using $\alpha_t = \text{softmax}(e)_t$. Finally, the sequence $\mathbf{Z}$ is aggregated into a single global context vector $\mathbf{c}$ via the weighted sum, emphasizing the most informative time steps (i.e., the peak of a gesture) while suppressing static signal at background as follows:
\begin{equation}
    \mathbf{c} = \sum_{t=1}^T \alpha_t \mathbf{z}_t.
\end{equation}

\subsection{Semantic Router}
The context vector $\mathbf{c}$ is passed to the semantic router, denoted as $G(\cdot)$, which is a Residual Gating Network (ResGateNet)~\cite{8578564} for distinguishing CSI patterns from different domains. The semantic router acts as a classifier that maps the signal context to a probability vector over the $N$ trained activity classification specialists, one for each domain, as follows:
\begin{equation}
    \mathbf{p} = \text{Softmax}(\text{G}(\mathbf{c})) \in \mathbb{R}^{N}.
\end{equation}
It then performs a single selection to identify the most appropriate specialist $k^*$ as follows:
\begin{equation}
    k^* = \operatorname*{arg\,max}_{j \in \{1, \dots, N\}} p_j.
\end{equation}
where$p_j$ is the $j$-th dimension of $\mathbf{p}$. 
\label{subsec:inference}
Afterwards, $E_{k^*}(\cdot)$ is activated while the other specialists remain inactive. This design is crucial for edge deployments of CSI-based HAR systems with limited computational resources.

The selected specialist further processes the shared features $\mathbf{Z}$ to capture finer dynamics associated with the activity classes and is passed to the classification head $\Psi_k(\cdot)$. The classification result is derived by picking the maximum probability over the activity classes:
\begin{equation}
    \hat{y} = \Psi_k(E_{k^*}(\mathbf{Z})).
\end{equation}

To efficiently train the proposed framework, we design a novel protocol that separately optimizes the shared feature extraction and activity classification specialists. The protocol also trains the semantic router iteratively over the domains. The decoupled protocol efficiently optimizes the model with limited data streaming in.

\section{Decoupled Continual Training Protocol}
\label{sec:protocol}
In the SAMoE-C framework, the shared backbone extracts coarse features from raw CSI tensors. In contrast, the specialist networks should capture finer-grained domain-specific variations and are activated according to the semantic router. Since the SAMoE-C framework targets at scenarios with sequential domains, waiting to update all parameters until complete domain datasets are collected would incur significant computational overhead and postpone the training process which contradicts the target scenarios.

To this end, we propose a staged training protocol that decouples the optimization of the shared backbone, specialist networks, and semantic router. It consists of three phases: 1) \textbf{Initial Domain Training} optimizes the shared feature extraction backbone with CSI data of $\mathcal{D}_1$, which is frozen subsequently to ensure feature stability. 2) \textbf{Incremental Learning}, which optimizes a dedicated specialist for a new domain without altering the shared backbone and specialists that have been previously trained. 3) \textbf{Router Update} calibrates the gating mechanism of Semantic Router for activating specialists in adaptation to CSI data. Note that the Initial Domain Training takes place only once, whilst Incremental Learning and Router Update are invoked whenever a new domain appears.
\vspace{-4mm}

\subsection{Initial Domain Training}
The shared backbone is trained on the raw CSI data of initial domain $\mathcal{D}_1$. To form an end-to-end optimization pipeline, the shared backbone is trained along with the first specialist $\mathcal{E}_1$, by minimizing the following loss function:

\begin{equation}
\mathcal{L}_{\text{init}}(\theta_{\Phi}, \theta_{\mathcal{E}_1}) = -\sum_{(\mathbf{X}, y) \in \mathcal{D}_1} \log \text{Softmax}( \mathcal{E}_1( \Phi(\mathbf{X}; \theta_{\Phi}); \theta_{\mathcal{E}_1}))_{y}
\end{equation}

\noindent where $\theta_{\Phi}$ and $\theta_{\mathcal{E}_1}$ are respectively trainable parameters of the shared backbone and specialist, and $\Phi(\mathbf{X}; \theta_{\Phi})$ represents the latent feature vectors extracted by the backbone. Once trained, both shared backbone and $\mathcal{E}_1$ are frozen thereafter.

\subsection{Incremental Learning}
\label{sec:incremental_learning}
The training protocol switches to Incremental Learning after the Initial Domain Training. Specifically, a dedicated specialist $\mathcal{E}_s$ and the semantic router are updated with the CSI data from the $s$-th domain.

\subsubsection{Specialist Training}
Specialist $\mathcal{E}_s$ is trained with a similar strategy adopted the Initial Domain Training, which optimizes the training parameter $\theta_{\mathcal{E}_{s}}$ by minimizing the following loss function, where $\theta_{\mathcal{E}_s}$ represents the trainable parameters of $\mathcal{E}_s$:

\begin{equation}
\label{eq:spectrain}
\mathcal{L}_{inc}(\theta_{\mathcal{E}_s}) = -\sum_{(\mathbf{X}, y) \in \mathcal{D}_s} \log \text{Softmax}( \mathcal{E}_s( \Phi(\mathbf{X}); \theta_{\mathcal{E}_s}))_{y}
\end{equation}

\subsection{Router Update}
Once the specialist $\mathcal{E}_s(\cdot)$ is trained in timestep $s$ its parameters are frozen and the training protocol proceeds to update the Semantic Router. The training data are exclusively associated with $\mathcal{D}_s$ with the same ground-truth specialist index $k=s$ for all samples. Updating the semantic router solely with data from $\mathcal{D}_s$ can overwrite existing decision boundaries and data samples are routed to an inappropriate specialist, producing inaccurate results.

To mitigate such problem requires addressing not only catastrophic forgetting, but also with respect to constrained storage quota on edge devices. We employ an experience replay cache buffer $\mathcal{B}$, which is a tiny fraction of historical domain training data. The buffer $\mathcal{B}$ is populated via a balanced sampling strategy where a fixed fraction of samples in each past domain. For the current training stage $s$, we construct the router training set $\mathcal{D}_{\text{mix}}$ by combining current domain data $\mathcal{D}_{s}$ and all samples in $\mathcal{B}$. The buffer will be updated after training.

The router's objective is to minimize loss over this composite dataset:
\begin{equation}
    \mathcal{L_s}(\theta_{{G}})= - \sum_{(\mathbf{X}, k) \in \mathbb{B} \cup D^{s}} \log \text{Softmax}( G (\mathbf{c}(\mathbf{X})) )_{k},
\end{equation}

\noindent where $\theta_G$ are the trainable router parameters, $\mathbf{c}(\mathbf{X})$ is the context vector derived from input $\mathbf{X}$ via the frozen backbone and attention module. To ensure the stability and efficient training scheduler, the router will not be involved in consequent training for other specialists.

\begin{table}[t!]
\caption{Model Architecture Implemented for SAMoE-C Framework}
\label{tab:structure}
\centering
\vspace{-2mm}
\begin{tabular}{llc}
\hline
\textbf{No.} & \textbf{Layer / Operation} & \textbf{Output Shape} \\
\hline
\multicolumn{3}{l}{\textbf{Part A: Shared Backbone (Layers 1-5)}} \\
1 & Input CSI Tensor $\mathbf{X}$ & $(3, 10, 114)$ \\
2 & Reshape for Conv1D & $(10, 3, 114)$ \\
3 & Stem Conv1D (7x7) \& MaxPool1D & $(10, 64, 29)$ \\
4 & Conv Block 1 (64 ch, stride=1) & $(10, 64, 29)$ \\
5 & Conv Block 2 (96 ch, stride=1) & $(10, 96, 29)$ \\
6 & Conv Block 3 (144 ch, stride=1) & $(10, 144, 29)$ \\
7 & Conv Block 4 (256 ch, stride=1) $\rightarrow \mathbf{Z}$ & $(10, 256, 29)$ \\
\hline
\multicolumn{3}{l}{\textbf{Part B: Attention-based Router}} \\
8a & Global Average Pool $\mathbf{Z}$ \& Reshape & $(10, 256)$ \\
8b & Temporal Mean Pooling $\rightarrow \mathbf{c}$ & $(256)$ \\
8c & ResGateNet (3 Blocks, 256-dim) $\rightarrow \mathbf{p}$ & $(4)$ \\
\hline
\multicolumn{3}{l}{\textbf{Part C: Activated Specialist (Layers 6-8 + RNN)}} \\
9a & Specialist Conv Block 5 (128 ch, stride=2) & $(10, 128, 15)$ \\
9b & Specialist Conv Block 6 (96 ch, stride=2) & $(10, 96, 8)$ \\
9c & Specialist Conv Block 7 (64 ch, stride=2) & $(10, 64, 4)$ \\
10 & AdaptiveAvgPool1D (Spatial) $\rightarrow$ Reshape & $(10, 64)$ \\
11 & Bi-GRU (2 Layers, hid=128) $\rightarrow \mathbf{z}_{j^*}$ & $(256)$ \\
12 & Classification Head (Linear) & $(27)$ \\
\hline
\multicolumn{3}{p{0.45\textwidth}}{\footnotesize Notation: Output shapes are shown per single sample instance. The Bi-GRU output dimension is 256 due to the concatenation of bidirectional hidden states.}
\end{tabular}
\vspace{-6mm}
\end{table}

\section{Experimental Evaluation}
\label{sec:experiments}
We conduct experiments to validate our proposed SAMoE-C framework targeted on the MM-Fi dataset~\cite{wen2023mmfi}, focusing on system performance and  computational efficiency.
\vspace{-4mm}

\subsection{Experimental Setup}
\textbf{Dataset:} We utilize the CSI data from the multi-modal MM-Fi dataset, which is extracted from WiFi signal collected by a TP-Link N750 router in four environments via the Atheros CSI Tool.  It features \(K=27\) human activities across four environments, including two living rooms (\(\mathcal{D}_1, \mathcal{D}_2\)) and two meeting rooms (\(\mathcal{D}_3, \mathcal{D}_4\)). By default, the domains appear with the order of \(\mathcal{D}_1 \to \mathcal{D}_2 \to \mathcal{D}_3 \to \mathcal{D}_4\). To assess the impact of the initial domain, we also consider the case with \(\mathcal{D}_3 \to \mathcal{D}_1 \to \mathcal{D}_2 \to \mathcal{D}_4\).

\textbf{Model and Hyper-parameter Configurations:} Following the system design from Section~\ref{sec:framework}, we implement an RNN-based network with attention-based router for experiment. The model architecture adopted in SAMoE-C is detailed in Table \ref{tab:structure}. Using the AdamW optimizer, the specialists are trained for 35 epochs, while the router is trained for 20 epochs. The batch size and learning rate are chosen as 64 and \(3\times10^{-4}\), respectively, which are applied consistently throughout the training of all components. In addition, we use \(\rho=0.05\) to construct the replay cache buffer for training the router.

\textbf{Baseline Schemes:} We consider two baseline CL schemes for comparison, including:
\begin{itemize}
    \item \textbf{Basic CL:} The Basic CL method utilizes a single specialist in SAMoE-C without the router. The model is re-trained with the loss function in (\ref{eq:spectrain}) when a new domain appears. While simple, this method is susceptible to catastrophic forgetting.
    \item \textbf{MEMO~\cite{zhou2023model}:} The MEMO baseline corresponds to the fully implemented multi-specialist SAMoE-C framework also without the router. The training process is the same as SAMoE-C specialists. During inference, all specialists will be activated to provide outputs. The final classification result is obtained by aggregating the outputs of all the specialists via a simple logit-based adapter.
\end{itemize}
\vspace{-5mm}

\begin{figure*}[t]
    \centering
    \begin{minipage}{0.5\textwidth}
        \centering
        \includegraphics[width=\linewidth]{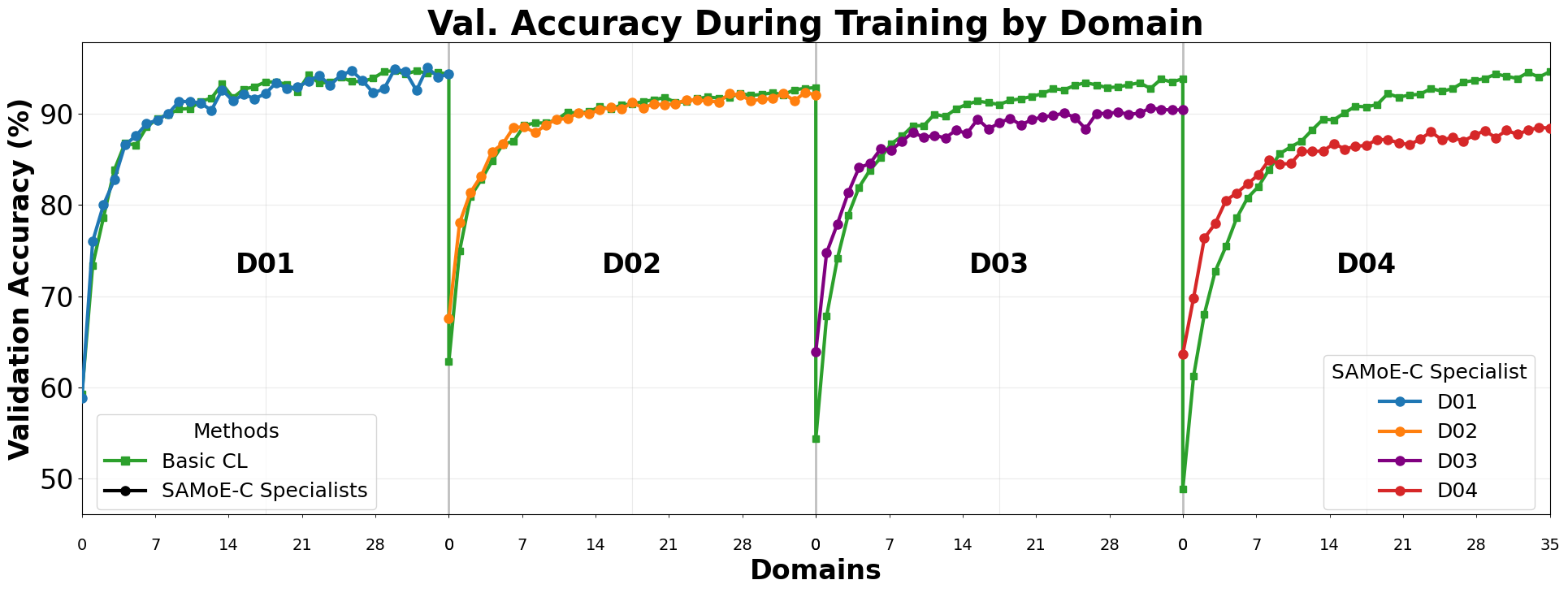} 
        \caption{Validation accuracy trends for Basic CL and Specialists across domains during \textbf{training}. The MEMO baseline shares the trajectory with SAMoE-C as they have the same Specialists.}
        \label{fig:curvesmain}
    \end{minipage}\hfill
    \begin{minipage}{0.45\textwidth}
        \centering
        \includegraphics[width=\linewidth]{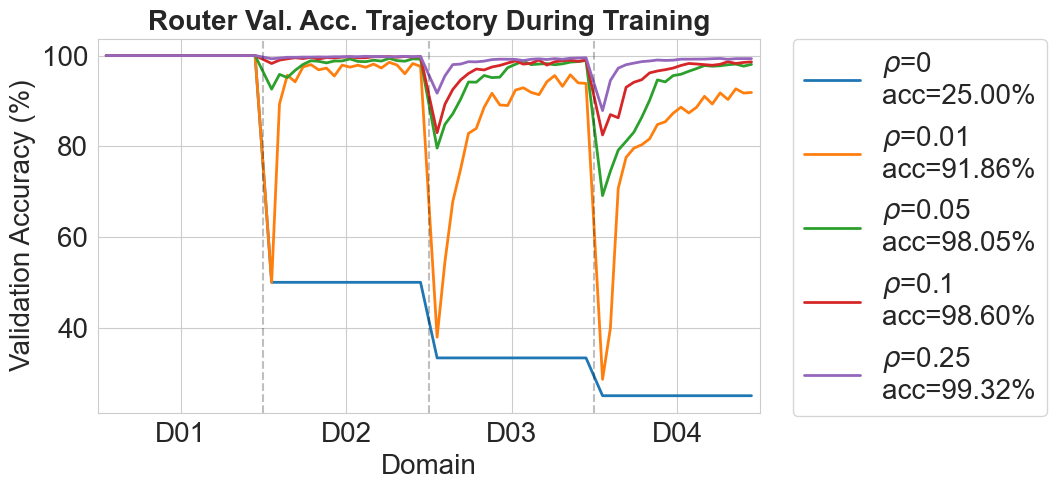} 
        \caption{Router's domain discrimination accuracy during training with varying replay buffer size ratio.}
        \label{fig:curvesrouter}
    \end{minipage}
\vspace{-6mm}
\end{figure*}

\begin{table}[htbp]
\centering
\caption{Progressive HAR Performance Comparison}
\vspace{-2mm}
\label{tab:cross_scene_comparison}
\setlength{\tabcolsep}{8pt}
\renewcommand{\arraystretch}{1.1}

\begin{tabular}{|c|c|c|c|}
\hline
\multicolumn{2}{|c|}{\textbf{Method}} & \textbf{Avg. Past Acc.} & \textbf{New Acc.} \\
\hline

\multirow{4}{*}{\shortstack[c]{\textbf{Basic CL}\\\textbf{Single Model Over Domains}}}
  & \(\mathcal{D}_1\)  & N/A    & 94.35\% \\
\cline{2-4}
  & \(\mathcal{D}_2\)  & 3.22\% & 92.85\% \\
\cline{2-4}
  & \(\mathcal{D}_3\)  & 3.65\% & 93.84\% \\
\cline{2-4}
  & \(\mathcal{D}_4\)  & 3.74\% & 94.60\% \\
\hline

\multirow{4}{*}{\shortstack[c]{\textbf{SAMoE-C}\\\textbf{Specialists}\\\textbf{Domain Specific}}}
  & \(\mathcal{D}_1\)  & \multicolumn{2}{c|}{94.33\%} \\
\cline{2-4}
  & \(\mathcal{D}_2\)  & \multicolumn{2}{c|}{92.07\%} \\
\cline{2-4}
  & \(\mathcal{D}_3\)  & \multicolumn{2}{c|}{90.41\%} \\
\cline{2-4}
  & \(\mathcal{D}_4\)  & \multicolumn{2}{c|}{88.42\%} \\
\hline

\end{tabular}

\vspace{-6mm}
\end{table}

\subsection{System Performance and Efficiency}
In this section, we evaluate the HAR capabilities and computational complexity of the proposed SAMoE-C framework. We first analyze the learning dynamics across domains. The effectiveness and efficiency of our design are then discussed. 

\subsubsection{Anti-Forgetting by Staged Analysis} We visualize the learning trajectory of different methods in Fig.~\ref{fig:curvesmain}, which are measured on the validation datasets of different domains. It is observed that while Basic CL exhibits slightly higher accuracy within individual domains, it suffers from catastrophic forgetting. According to Table~\ref{tab:cross_scene_comparison}, measured on a balanced mixture of domain being progressively added, the average accuracy collapses to 3.74\%. In contrast, the SAMoE-C specialists overcomes this issue attributed to our training protocol, i.e., when adapting a new environments, by freezing previously trained shared backbone and decoupling HAR problem into domain-specific classification tasks. The accuracy of SAMoE-C specialists decline gracefully from 94.33\% on \(\mathcal{D}_1\) to 88.42\% on \(\mathcal{D}_4\).
\vspace{-4mm}

\begin{table}[ht]
    \centering
    \caption{Final Average HAR Accuracy and Inference Efficiency}
    \label{tab:final_comparison}
    \setlength{\tabcolsep}{4pt}
    \begin{tabular}{@{}lcc@{}}
        \toprule
            \textbf{Method} & \parbox[t]{2.2cm}{\centering \textbf{Final Avg. HAR Acc. (\%)}} & \parbox[t]{2.5cm}{\centering \textbf{Inference Cost (MFLOPS/Sample)}} \\
        \midrule
        Basic CL & 29.56 & 198.4 \\
        MEMO & 87.69 & 797.8 \\
        \textbf{SAMoE-C} & \textbf{81.66} & \textbf{199.1} \\
        SAMoE-C(Alt. Order) & 81.24 & 199.1 \\
        \bottomrule
    \end{tabular}
\vspace{2mm}
{\footnotesize

Note: Measured on balanced test set. Samples in (3, 114, 10) shape.}
\end{table}

\subsubsection{Overall HAR Accuracy and Computational Efficiency} The domain specialists in SAMoE-C are orchestrated by the semantic router. Fig.~\ref{fig:curvesrouter} depicts the domain discrimination accuracy of the semantic router. Under the default replay buffer ratio of $\rho=0.05$, the accuracy approaches 98.05\% after iteratively training on all domains. Guided by the highly accurate domain discrimination results, the SAMoE-C framework yields a competitive average HAR accuracy of 81.66\% as reported in Table~\ref{tab:final_comparison}, measured on a balanced test set. Additionally, SAMoE-C maintains a consistent 81.24\% under an alternative domain appearance order, i.e., \(\mathcal{D}_3 \to \mathcal{D}_1 \to \mathcal{D}_2 \to \mathcal{D}_4\), proving its resilience to initial domain variations.

Notably, although the MEMO baseline achieves a higher accuracy of 87.69\%, it incurs a linearly scaling inference cost of 797.8 MFLOPS per sample. By utilizing the semantic router to selectively activate a single specialist, SAMoE-C drastically reduces the computational demand to 199.1 MFLOPS per sample. Moreover, although it needs a negligible 0.7 MFLOPS routing overhead compared to the Basic CL baseline, a significant 52.10\% accuracy improvement is achieved. Consequently, SAMoE-C delivers a favorable accuracy-efficiency trade-off, making it more suitable for resource-constrained edge deployments.

\subsubsection{Ablation Study on Replay Buffer Size}
To investigate the impact of the replay buffer size, we also conduct an ablation study by varying \(\rho\) in Fig.~\ref{fig:curvesrouter}. When \(\rho = 0\), i.e., without historical data for training the router, severe catastrophic forgetting is observed with a routing accuracy of only 25.00\%. This is primarily due to a strong bias towards the \(\mathcal{D}_4\), which underscores the necessity of the replay buffer. Interestingly, introducing a minimal replay buffer with \(\rho = 0.01\) boosts the routing accuracy to \textbf{91.86\%}. While expanding the buffer size further enhances the routing accuracy, e.g., \textbf{99.32\%} at \(\rho = 0.25\), the gain is marginal. In practice, \(\rho = 0.05\) can be used to balance accuracy and buffer size demand on edge devices.
\vspace{-2mm}

\section{Conclusion}
\label{sec:conclusion}
In this letter, we propose the SAMoE-C framework for robust cross-domain CSI-based HAR. By integrating an MoE architecture and a novel training protocol, our method successfully preserves knowledge from previously seen environments while adapting to new ones, achieving high domain discrimination and activity classification performance, offering a practical accuracy-efficiency tradeoff with constant inference cost. Further research will keep improving the framework by pretraining the backbone on a more comprehensive dataset and exploring soft routing for specialist activation.
\vspace{-3mm}
\bibliographystyle{IEEEtran}
\bibliography{references} 

@ARTICLE{10836707,
  author={Zhu, Tianyun and Dong, Yilin and Zhou, Yong and Zhu, Changming and Cao, Lei},
  journal={IEEE JBHI}, 
  title={Cross-Domain Human Activity Recognition via Domain Adaptation and Fused Attention}, 
  year={2025},
  volume={29},
  number={8},
  pages={5394-5404},
  doi={10.1109/JBHI.2025.3528008}}

@article{10.1145/3530682,
author = {Xiao, Jiang and Li, Huichuwu and Wu, Minrui and Jin, Hai and Deen, M. Jamal and Cao, Jiannong},
title = {A Survey on Wireless Device-free Human Sensing: Application Scenarios, Current Solutions, and Open Issues},
year = {2022},
issue_date = {May 2023},
publisher = {Association for Computing Machinery},
address = {New York, NY, USA},
volume = {55},
number = {5},
issn = {0360-0300},
journal = {ACM Comput. Surv.},
month = dec,
articleno = {88},
numpages = {35}
}

@inproceedings{rebuffi2017icarl,
  title={{iCaRL}: {I}ncremental classifier and representation learning},
  author={Rebuffi, Sylvestre-Alvise and Kolesnikov, Alexander and Sperl, Georg and Lampert, Christoph H},
  booktitle={Proc. IEEE Conf. Comput. Vis. Pattern Recognit. (CVPR)},
  address={Honolulu, HI, USA},
  year={2017},
  doi={10.1109/CVPR.2017.587}
}

@inproceedings{zhou2023model,
  title={A model or 603 exemplars: {T}owards memory-efficient class-incremental learning},
  author={Zhou, Da-Wei and Wang, Qi-Wei and Ye, Han-Jia and Zhan, De-Chuan},
  booktitle={Proc. Int. Conf. Learn. Represent. (ICLR)},
  year={2023}
}

@article{wen2023mmfi,
  author={Wen, Han and Zhang, Jianyuan and Zhang, Dazhou and Yang, Xing-Dong},
  journal={IEEE Internet Things J.}, 
  title={{M}m-{F}i: {A} {M}ulti-{M}odal {H}uman-{C}omputer {I}nteraction {P}latform {W}ith {W}i-{F}i}, 
  year={2023},
  volume={10},
  number={15},
  pages={13329-13342},
  doi={10.1109/JIOT.2023.3262024}
}

@inproceedings{shazeer2017outrageously,
  title={{O}utrageously large neural networks: {T}he sparsely-gated mixture-of-experts layer},
  author={Shazeer, Noam and Mirhoseini, Azalia and Maziarz, Karol and Davis, Andy and Le, Quoc and Hinton, Geoffrey and Dean, Jeff},
  booktitle={Proc. Int. Conf. Learn. Represent. (ICLR)},
  year={2017}
}

@article{bolcskei2006mimo,
  title={{MIMO-OFDM} wireless systems: basics, perspectives, and challenges},
  author={B{\"o}lcskei, Helmut and Gesbert, David and Paulraj, Arogyaswami J},
  journal={IEEE Wireless Commun.},
  volume={13},
  number={4},
  pages={31--37},
  year={2006}
}

@inproceedings{10.5555/3692070.3694593,
  author = {Zhao, Xuyang and Wang, Huiyuan and Huang, Weiran and Lin, Wei},
  title = {A statistical theory of regularization-based continual learning},
  year = {2024},
  booktitle = {Proc. Int. Conf. Mach. Learn. (ICML)}
}

@article{dhekane2024transfer,
  title={Transfer {L}earning in {H}uman {A}ctivity {R}ecognition: {A} {S}urvey},
  author={Dhekane, Sourish Gunesh and Ploetz, Thomas},
  journal={arXiv preprint arXiv:2401.10185},
  year={2024}
}

@article{cook2013transfer,
  title={Transfer learning for activity recognition: a survey},
  author={Cook, Diane and Feuz, Kyle D and Krishnan, Narayanan C},
  journal={Knowl. Inf. Syst.},
  volume={36},
  pages={537--556},
  year={2013}
}

@article{ganin2016domain,
  title={Domain-adversarial training of neural networks},
  author={Ganin, Yaroslav and Ustinova, Evgeniya and Ajakan, Hana and Germain, Pascal and Larochelle, Hugo and Laviolette, Fran{\c{c}}ois and Marchand, Mario and Lempitsky, Victor},
  journal={J. Mach. Learn. Res.},
  volume={17},
  number={59},
  pages={1--35},
  year={2016}
}

@ARTICLE{10476327,
  author={Ratnam, Vishnu V. and Chen, Hao and Chang, Hao-Hsuan and Sehgal, Abhishek and Zhang, Jianzhong},
  journal={IEEE Transactions on Wireless Communications}, 
  title={Optimal Preprocessing of WiFi CSI for Sensing Applications}, 
  year={2024},
  volume={23},
  number={9},
  pages={10820-10833},
  doi={10.1109/TWC.2024.3376332}}

@INPROCEEDINGS{8578564,
  author={Yu, Xin and Yu, Zhiding and Ramalingam, Srikumar},
  booktitle={2018 IEEE/CVF Conference on Computer Vision and Pattern Recognition}, 
  title={Learning Strict Identity Mappings in Deep Residual Networks}, 
  year={2018},
  volume={},
  number={},
  pages={4432-4440},
  doi={10.1109/CVPR.2018.00466}}

\end{document}